%% file: main.tex
\documentclass{article}
\usepackage{iclr2027_conference,times}
\input{math_commands.tex}

\usepackage{hyperref}
\hypersetup{hidelinks,pdfauthor={Yonggang Li, Boqian Han, Rui SHENG, Yushi Sun, Zixin CHEN},pdftitle={RegulaTutor: A Theory-Informed Framework for Long-Term Personalized Learning}}
\usepackage{url}
\usepackage{booktabs}
\usepackage{tabularx}
\usepackage{flafter}
\usepackage{placeins}
\usepackage{float}
\usepackage{tikz}
\usetikzlibrary{arrows.meta,positioning}
\title{RegulaTutor: A Theory-Informed Framework for Long-Term Personalized Learning}
\author{\makebox[\textwidth][c]{\normalfont\bfseries
Yonggang Li$^{1,}$\thanks{These authors contributed equally to this work.},\enspace
Boqian Han$^{2,*}$,\enspace Rui SHENG$^{1}$,\enspace Yushi Sun$^{1}$,\enspace Zixin CHEN$^{1}$}\\[0.5em]
\makebox[\textwidth][c]{\normalfont\small $^{1}$The Hong Kong University of Science and Technology}\\[0.15em]
\makebox[\textwidth][c]{\normalfont\small $^{2}$The Chinese University of Hong Kong}%
}
\iclrfinalcopy
\begin{document}
\maketitle
\lhead{Preprint}
\begin{abstract}
Long-term personalized tutoring requires language models to translate a learner's evolving history into appropriate instructional support. Existing LLM-based tutoring systems use memory and retrieval to personalize responses, but the pedagogical rationale linking historical evidence to teaching strategies is not always explicit. We introduce \textit{RegulaTutor}, a theory-informed framework for long-term personalized tutoring that connects historical evidence, learner interpretation, and learner-facing instruction. Guided by self-regulated learning (SRL) research, its tutoring harness operationalizes a cognitive, affective, metacognitive, and motivational (CAMM) framework for deriving evidence-supported learner interpretations and teaching policies. The resulting policies retain links to their supporting evidence and are updated as new observations arrive. This staged workflow also supports systematic refinement through \textit{PairRoute}, an offline procedure that uses intermediate execution records to distinguish harness, model, and mixed failures.
\textit{PairRoute} proposes and evaluates targeted harness revisions, then constructs reviewed stage-level supervision for confirmed model failures remaining under the selected harness. We focus this refinement on historical evidence acquisition, which underpins personalized instruction and allows outputs to be checked against learning records and reference answers. 
RegulaTutor improves historical-evidence accuracy, learner diagnosis, and judged teaching quality on average across evaluated model configurations. PairRoute results show an endpoint improvement in standalone-model historical-evidence accuracy over the refinement trajectory.
Together, these results highlight \textit{RegulaTutor}'s potential to support lifelong personalized education through instructional policies that evolve with learner histories and targeted improvements to the tutoring system.
\end{abstract}
\input{figures/regulatutor_intro_overview}
\input{sections/introduction}

\input{sections/method}

\input{sections/experiments}

\input{sections/related_work}
\input{sections/discussion}

\input{sections/conclusion}

\label{maintext:end}
\clearpage
\input{sections/statements}
\bibliography{references}
\bibliographystyle{iclr2027_conference}
\clearpage
\appendix
\input{sections/appendix}
\end{document}

%% file: math_commands.tex
\usepackage{amsmath,amsfonts,bm}

\def\eqref#1{equation~\ref{#1}}

\def\1{\bm{1}}

\DeclareMathAlphabet{\mathsfit}{\encodingdefault}{\sfdefault}{m}{sl}
\SetMathAlphabet{\mathsfit}{bold}{\encodingdefault}{\sfdefault}{bx}{n}

%% file: figures/regulatutor_intro_overview.tex
\begin{figure}[!t]
  \centering
  \includegraphics[width=\textwidth]{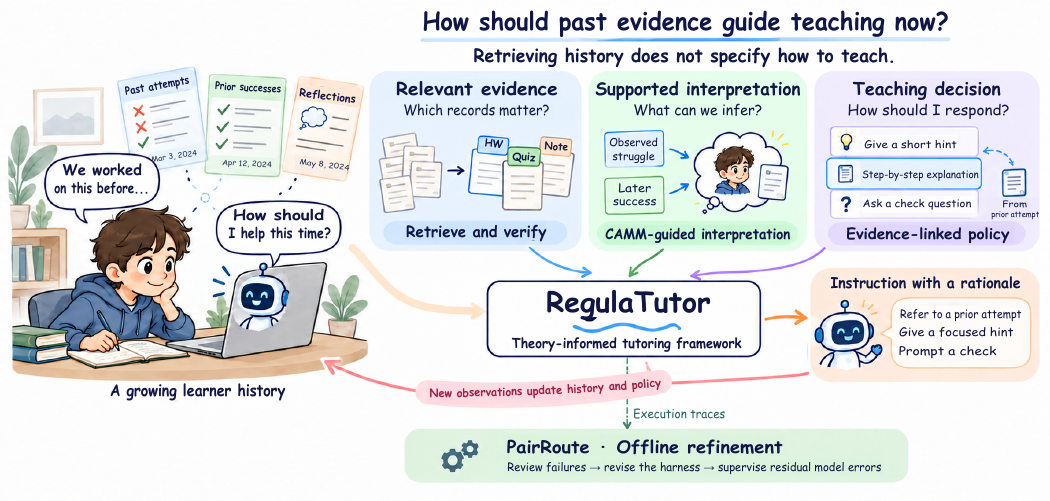}
  \caption{\textbf{RegulaTutor connecting learner history to instructional decisions.} The learner–tutor setting (left) motivates three questions (top): which records are relevant, what interpretations they support, and how those interpretations should guide teaching. RegulaTutor makes these connections explicit through CAMM-guided learner interpretation and evidence-linked teaching policies, with learner history and policies updated as new observations arrive. PairRoute (bottom) uses execution traces to guide offline harness revisions and construct reviewed stage-level supervision for residual model errors.}
  \label{fig:framework-overview}
\end{figure}

%% file: sections/introduction.tex
\section{Introduction}
\label{sec:introduction}
Large language models (LLMs) are increasingly being developed into tutors that explain concepts, provide feedback, and personalize instruction \citep{zhaoDeepTutorAgenticPersonalized2026,perczelTeachLMPostTrainingLLMs2025}.
Yet learning unfolds across repeated attempts and interactions: a learner's difficulties may persist, recur, or change as understanding develops.
Long-term personalized tutoring must therefore relate the current task to accumulated learner history and adapt as new observations arrive.
However, recent evaluations highlight the difficulty of translating long-term learning histories into accurate diagnosis and appropriate instructional decisions \citep{liLongTutorBenchmarkingLarge2026}.

Existing LLM-based tutoring systems use memory, retrieval, and learner profiles to support history-based personalization \citep{zhaoDeepTutorAgenticPersonalized2026,wuTeachingAccordingStudents2025}.
However, retrieving relevant records does not by itself specify how those records should guide teaching \citep{cohnTheoryAdaptiveScaffolding2026,guresTailoringScaffoldingDiagnostic2026}. Repeated errors, for example, identify a difficulty worth investigating, but do not by themselves determine whether the tutor should reteach a concept, scaffold a procedure, or prompt the learner to inspect their own reasoning. Choosing an appropriate response therefore requires interpreting such evidence in terms of the learner's knowledge and learning processes, and aligning instructional support with that interpretation \citep{guresTailoringScaffoldingDiagnostic2026}. This motivates an explicit, theory-informed connection from historical evidence to learner interpretation and instructional action.

We introduce \textit{RegulaTutor}, a theory-informed framework for long-term personalized tutoring that connects historical evidence, learner interpretation, and learner-facing instruction (Figure~\ref{fig:framework-overview}).
Guided by self-regulated learning (SRL) research, its tutoring harness draws on cognitive, affective, metacognitive, and motivational (CAMM) processes to derive evidence-supported learner interpretations \citep{molenaarMeasuringSelfregulatedLearning2023,azevedoLessonsLearnedFuture2022}.
These interpretations inform learner-specific teaching policies that specify the support mode, explanation depth, references to prior learning episodes, and metacognitive or motivational support.
The harness translates these policies into instructional directives while retaining links to the supporting records.

Across interactions, the harness maintains learner history and updates these policies as new observations become available. Its shared system prompt also encourages learner goal setting, self-checking, and reflection when appropriate.
Together, these mechanisms provide an inspectable path from a learner's accumulated history to the tutor's instructional choices.

Making these intermediate stages explicit also supports improvement of the tutoring system itself.
An incorrect final response alone does not reveal whether the failure comes from the harness's access to or representation of evidence, the model's execution of a feasible stage, or both.
RegulaTutor records intermediate model inputs and outputs so that failures can be examined at the stages where they occur.
To guide targeted improvements, the framework includes \textit{PairRoute}, an offline refinement procedure building on trajectory-informed harness optimization and model--harness co-evolution \citep{leeMetaHarnessEndtoEndOptimization2026,chenHarnessForgeJointHarness2026,chenCoHarnessCoEvolvingHarnesses2026}.
PairRoute uses these execution records to diagnose harness, model, and mixed failures, evaluate targeted harness revisions, and construct reviewed stage-level supervision for model failures that remain under the selected harness.

We focus this refinement study on historical evidence acquisition, which supplies the factual basis for personalized instruction and provides targets that can be checked against learning records and reference answers \citep{liLongTutorBenchmarkingLarge2026}.
RegulaTutor thus supports adaptation at two timescales: new learner observations inform interpretations and teaching policies during tutoring, while execution failures guide offline refinement of the harness and model.

We evaluate \textit{RegulaTutor} on \textsc{LongTutor}, a benchmark for tutoring over long learning histories \citep{liLongTutorBenchmarkingLarge2026}. Fixed-backbone comparisons and component analyses examine the contribution of the harness and its instructional mechanisms.
Across the evaluated model configurations, RegulaTutor improves historical evidence acquisition and learner diagnosis on average.
Our focused refinement study uses Qwen3.5-4B to examine successive model checkpoints under different harness versions.
PairRoute's development-set trajectory shows gains in evidence processing under the selected harness; crossed model--harness evaluations further characterize how model updates interact with harness revisions.

Our contributions are threefold:
\begin{itemize}
    \item We introduce \textit{RegulaTutor}, a theory-informed tutoring framework that connects historical evidence, learner interpretations, and instructional directives, while updating learner history and tutoring policies across interactions.
    \item We develop \textit{PairRoute}, an offline refinement procedure within \textit{RegulaTutor} that uses intermediate execution records to diagnose failures, guide harness revisions, and construct reviewed stage-level supervision for model failures remaining under the selected harness.
    \item We evaluate \textit{RegulaTutor} on long-history tutoring, examining evidence acquisition, diagnosis, and teaching. A focused study with Qwen3.5-4B further examines how harness revisions and model updates affect tutoring performance.

\end{itemize}

%% file: sections/method.tex
\section{Method}
\label{sec:method}
\textit{RegulaTutor} connects historical evidence to learner interpretations and instructional decisions through cognitive, affective, metacognitive, and motivational (CAMM) processes. We further develop \textit{PairRoute} to use execution traces for harness revision and stage-level model supervision. Figure~\ref{fig:regulatutor-two-timescales} summarizes both workflows.
We first describe the tutoring harness, then introduce \textit{PairRoute}, which uses its execution records to guide harness revisions and stage-level model supervision.

\input{figures/regulatutor_two_timescales}

\subsection{From Learner History to Instructional Decisions}
\label{sec:harness}

At interaction $i$ with learner $u$, RegulaTutor uses the current task $q_{u,i}$ and history $\mathcal{L}_{u,i}$ to derive a learner interpretation, update tutoring policy $P_{u,i}$, and generate a response.
Model weights and shared harness rules remain unchanged during tutoring; learner history and policy evolve across interactions.

\paragraph{Learner history.}
RegulaTutor accumulates observations, dialogue, and policy state during ongoing sessions, or accepts a self-contained history for retrospective tasks.
From this history, it computes learner statistics $S_{u,i}$ and a bounded retrieval view $\mathcal{B}_{u,i}$, separating accumulated learner information from the context supplied to individual model calls.

\paragraph{Historical evidence acquisition.}
The Evidence Resolution Loop (ERL) identifies the information needed for $q_{u,i}$. For questions about a particular attempt, it retrieves focused records; for questions spanning multiple records, mapper calls select observations from history chunks for consolidation. Resolver calls combine the evidence into answers with source identifiers. Citation checks and bounded retrieval expansion address unresolved information needs. We write
\begin{equation}
(E_{u,i},A_{u,i})=\operatorname{ERL}_{M,H}(\mathcal{B}_{u,i},q_{u,i}),
\label{eq:erl}
\end{equation}
where $M$ and $H$ denote the language model and tutoring harness, respectively, $E_{u,i}$ preserves source evidence, and $A_{u,i}$ contains per-need answers and support status.
These outputs provide the evidence for the subsequent learner interpretation.

\paragraph{CAMM-guided learner interpretation.}
We draw on self-regulated learning (SRL) and its Metacognitive and Affective Model (MASRL) to connect observations with task-specific support \citep{zimmerman2002,efklides2011}, using CAMM as the organizing framework \citep{molenaarMeasuringSelfregulatedLearning2023}. RegulaTutor constructs the current learner interpretation $C_{u,i}$ from admitted evidence and statistics of the corresponding source events. In the task-verified configuration, this evidence consists of records cited by ERL.. Each tentative learner claim retains its supporting evidence, uncertainty, instructional implications, and unsupported inferences to avoid.

\paragraph{Evidence-linked teaching policies.}
Our guidelines map supported interpretations to policy proposals $\mathcal{A}_{u,i}$ for support mode, explanation depth, history references, and metacognitive or motivational support (Table~\ref{tab:app-camm} in Appendix~\ref{app:camm}). For example, repeated errors on a concept support an interpretation of local difficulty, leading to stepwise explanation and a focused check grounded in a prior attempt.
These errors do not by themselves establish low motivation. RegulaTutor updates the previous policy by screening proposals for evidence support and confidence, penalizing unnecessary switching, and resolving competing assignments to the same field. It retains policy provenance and re-examines learner interpretations as new evidence arrives. Appendix~\ref{app:camm} gives the update rules and representation details.

\paragraph{Instructional directives and response generation.}
To translate the interpretation and policy into learner-facing instruction, RegulaTutor combines $C_{u,i}$ and $P_{u,i}$ into instructional directives $D_{u,i}$. It then constructs the input for the final response-generation call:
\begin{equation}
\begin{split}
x^{\mathrm{final}}_{u,i}&=\operatorname{Prompt}(q_{u,i},S_{u,i},E_{u,i},A_{u,i},P_{u,i},D_{u,i},\mathcal{D}_{u,i}),\\
y_{u,i}&\sim M(\cdot\mid x^{\mathrm{final}}_{u,i}).
\end{split}
\label{eq:prompt}
\end{equation}
Recent dialogue $\mathcal{D}_{u,i}$ preserves speaker roles to support follow-up responses. A shared system prompt encourages task-appropriate goal setting, self-checking, and reflection. RegulaTutor validates output structure and records intermediate execution, providing the traces used for offline refinement.

\subsection{PairRoute: Trajectory-Guided Offline System Refinement}
\label{sec:pairroute}
PairRoute uses RegulaTutor's execution records to guide updates targeting historical evidence acquisition. At offline round $t$, it operates on model-harness pair $(M_t,H_t)$ through two reviewing-agent roles: Router1 proposes and evaluates harness revisions; Router2 reviews residual failures and prepares model supervision under the selected harness. Selecting the harness before preparing supervision allows Router2 to assess failures against the inputs and requirements of the selected harness \citep{chenHarnessForgeJointHarness2026,chenCoHarnessCoEvolvingHarnesses2026}.

\paragraph{Trajectory-based failure diagnosis.}
Both router agents examine the task, reference evaluation, and actual stage inputs and outputs. A harness diagnosis requires evidence of defective access, representation, or call requirements; a model diagnosis requires sufficient visible support and a feasible local task that the model fails to execute. Mixed cases involve both. The routers separate source or evaluation problems from execution failures and defer unresolved cases. Software validates call identities and evidence pointers supporting each judgment (Appendix~\ref{app:routing}).

\paragraph{Router1: harness revision.}
Router1 uses training discovery trajectories to propose bounded revisions addressing harness and mixed failures. It also examines successes to specify behavior to preserve, then evaluates candidates with $M_t$ fixed using static checks, targeted controls, and full-task probes. These evaluations inform selection of $H_{t+1}$ under the round's recorded protocol, which may retain $H_t$.

\paragraph{Router2: supervision preparation.}
Router2 examines execution records under $(M_t,H_{t+1})$ and diagnoses failures again, because harness changes can alter the evidence available to a call and invalidate earlier failure assignments. For confirmed model failures, it localizes eligible calls, obtains teacher targets, and reviews them against the actual input, evidence, and output requirements. Only supported corrections enter training; other cases remain outside the correction set.

\paragraph{Stage-level training and update evaluation.}
Each stage-level training example consists of a recorded model-call input and a reviewed target. We combine reviewed corrections with reviewed correct current-model outputs, including correct stages within failed tasks, and apply completion-only supervised fine-tuning to obtain $M_{t+1}$.
Under the selected harness $H_{t+1}$, we compare $M_{t+1}$ with its parent model $M_t$ on evidence acquisition, diagnosis, and teaching.
Appendix~\ref{app:training} details training, and Appendix~\ref{app:versions} records the round-level update decisions.

%% file: figures/regulatutor_two_timescales.tex
\begin{figure}[t]
  \centering
  \includegraphics[width=\linewidth]{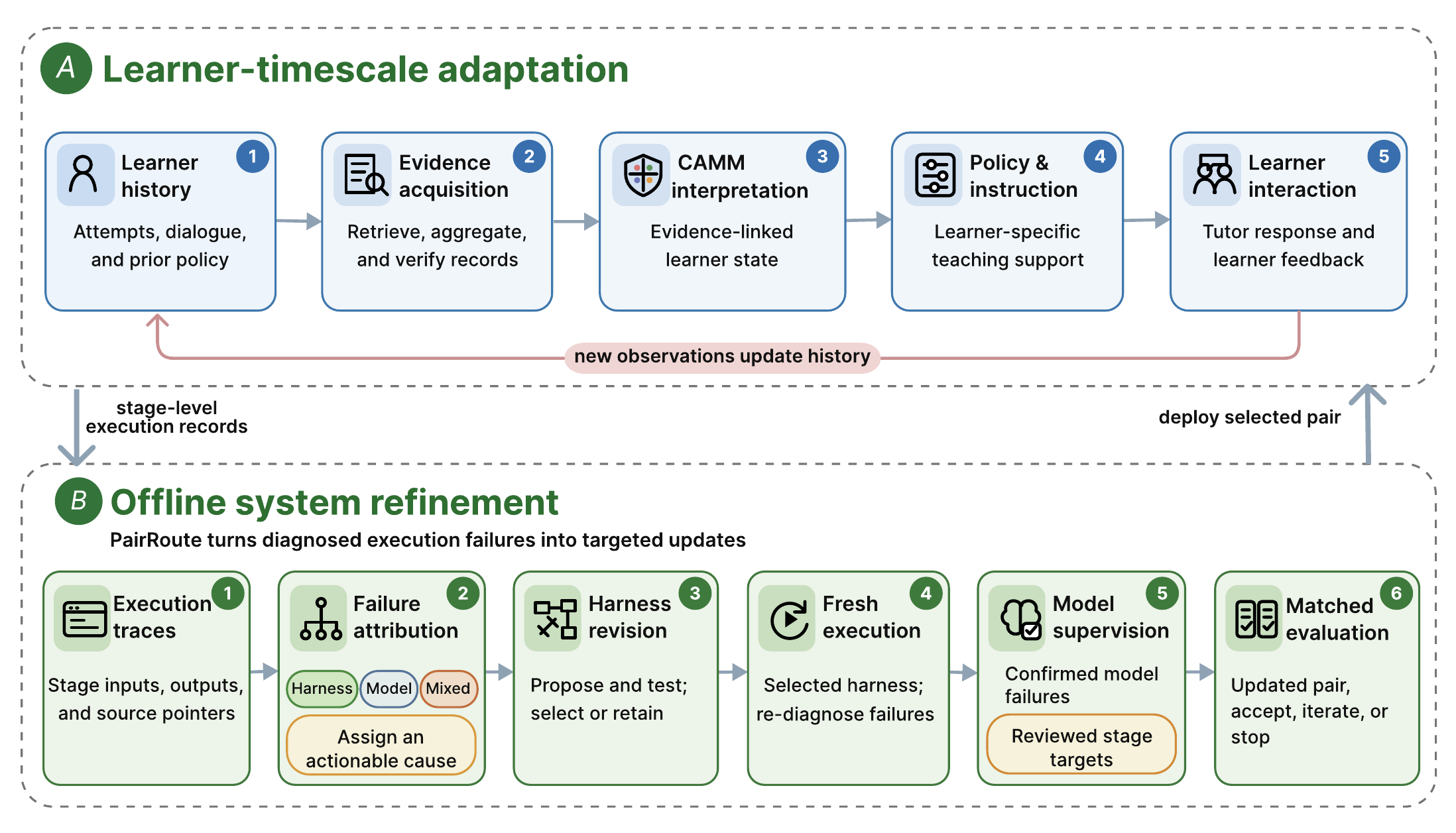}
  \caption{RegulaTutor adapts at two timescales. During tutoring, accumulated learner history supports evidence acquisition, CAMM-guided interpretation, and learner-specific instructional policies that update with new interactions. Offline, PairRoute attributes execution failures to the harness, model, or both and routes them to targeted harness revision or reviewed model supervision. The updated model--harness pair is evaluated before the next round or deployment.}
  \label{fig:regulatutor-two-timescales}
\end{figure}

%% file: sections/experiments.tex
\section{Experiments}
\label{sec:experiments}
We use LongTutor to evaluate how RegulaTutor acquires historical evidence, diagnoses learner difficulties, and supports instruction from long learner histories. We first compare each backbone with and without RegulaTutor, then examine the contributions of its components, and finally study PairRoute through successive model checkpoints and their crossed evaluations with harness versions.
\subsection{Experimental Setup}
\label{sec:exp_setup}

\paragraph{Benchmarks.}
We evaluate on \textsc{LongTutor}'s 1,000-case human-annotated split, which tests historical evidence acquisition, learner diagnosis, and teaching using 199-record learner histories \citep{liLongTutorBenchmarkingLarge2026}. Appendix~\ref{app:benchmark_protocols} specifies the metrics and judges.

\paragraph{Implementation details.}
We instantiate RegulaTutor with Qwen3.5-4B, DeepSeek-v4-pro, Gemini-3.7-flash, GPT-5.6-sol, Kimi-k3, and DeepSeek-v4.1-flash. All fixed-backbone comparisons use the final harness $H_1$ selected through PairRoute-guided refinement (Table~\ref{tab:longtutor-main}).

PairRoute starts from the initial Qwen3.5-4B model $M_0$ and harness $H_0$. Three rounds of supervised refinement on reviewed stage-level trajectories yield model checkpoints $M_1$--$M_3$; the selected harness $H_1$ is retained in later rounds, so $H_2=H_3=H_1$. Full model identities and decoding settings are given in Appendix~\ref{app:benchmark_protocols}. Data construction, training settings, and checkpoint identities appear in Appendices~\ref{app:training} and~\ref{app:versions}.

\paragraph{Baselines.}
We compare each model and checkpoint with and without RegulaTutor. We further compare with DeepTutor 1.5.1, a tutoring-harness baseline on DeepSeek-v4-pro \citep{zhaoDeepTutorAgenticPersonalized2026}. For GPT-5.6-sol, we also evaluate the native Codex agent as a third harness configuration. To examine PairRoute, we evaluate the crossed combinations of $M_0$--$M_3$ with $H_0$ and $H_1$.

\input{figures/table_longtutor_main}

\subsection{Main Results}
\label{sec:exp_main_results}

\label{sec:exp_harness}

\paragraph{Long-history tutoring.}
We compare each model and checkpoint with and without the final RegulaTutor harness in Table~\ref{tab:longtutor-main}.

Across all nine fixed-backbone comparisons, RegulaTutor improves Evidence accuracy by 13.2 percentage points and Diagnosis accuracy by 11.6 points on average. Table~\ref{tab:longtutor-main} reports these descriptive means separately for four local Qwen checkpoints and five API backbones. Six configurations show reliable Evidence gains, ranging from 10.9 points for DeepSeek-v4-pro (95\% CI [8.7, 13.0]) to 36.2 points for the initial Qwen checkpoint (95\% CI [34.1, 38.2]). These intervals quantify uncertainty in the estimated gains. For example, for DeepSeek-v4-pro, the plausible mean gain is 8.7--13.0 points. Both intervals exclude zero, indicating that the gains are statistically distinguishable from zero.

Across the four Qwen checkpoints, RegulaTutor improves Evidence accuracy by 15.5--36.2 percentage points and Diagnosis accuracy by 18.3--24.1 points. All corresponding 95\% confidence intervals exclude zero. At $M_0$, the Evidence gain is 36.2 points (95\% CI [34.1, 38.2]), with improvements on 738 cases, regressions on 48, and ties on 214. Diagnosis improves by 21.1 points (95\% CI [17.2, 24.9]), and Teaching rises by 0.132 points (95\% CI [.064, .200]).

The API results show the same Evidence--Teaching combination on two backbones: DeepSeek-v4-pro gains 10.9 Evidence points and 0.220 Teaching points, and Gemini gains 18.2 Evidence points and 0.506 Teaching points. Kimi-k3 gains 7.7 Diagnosis points and 0.698 Teaching points. These paired gains have intervals excluding zero.

\FloatBarrier

\subsection{Ablation Studies}
\label{sec:exp_ablation}

We evaluate component ablations on LongTutor with the ablation-batch reference held fixed. Removing CAMM interpretation lowers Evidence by 2.0 points and Teaching by 0.164, while fixing the policy lowers them by 1.5 points and 0.144 (Table~\ref{tab:harness-ablation}). These within-batch contrasts support the contribution of evidence-grounded interpretation and policy adaptation to the tutoring pathway. The absolute reference score is batch-specific and is therefore used only for paired contrasts.
\input{figures/table_harness_ablation}
\FloatBarrier

\subsection{Trace-Guided Refinement and Compatibility}
\label{sec:exp_coevolution}

\paragraph{Refinement protocol.}
PairRoute uses stage-level execution records to distinguish failures in evidence access or presentation from failures in the model's use of available evidence. For harness-related failures, it proposes focused revisions and tests each candidate while holding the model fixed. Under the selected harness, it reviews remaining model failures and correct executions to construct stage-level training examples.

We roll out the current model--harness pair on 369 discovery cases and analyze its execution trajectories to identify harness failures and model errors. A separate 32-case probe set screens candidate harness revisions, while a disjoint 100-case development set checks the selected configurations across refinement rounds. Only reviewed stage-level examples from the discovery cases enter model training.

Appendix~\ref{app:training} details the review and training procedure. The present study asks whether this supervision improves historical-evidence processing and how the resulting checkpoints behave under different harness versions.

\paragraph{Development-set results.}
On the 100-case development set, Evidence accuracy under the selected $H_1$ rises in every model-training round, from 212/300 (70.7\%) with $M_0$ to 250/300 (83.3\%) with $M_3$ (Table~\ref{tab:pairroute-development}). The archived $(M_0,H_0)$ run provides an initial-pair reference on the same cases. Later harness proposals were not retained, so $H_2=H_3=H_1$.
The first model update gains 6.0 percentage points under $H_1$, with a paired case-bootstrap 95\% interval of [1.7, 10.3]. 

\input{figures/table_pairroute_development}
\FloatBarrier

\paragraph{LongTutor Gold1000 results.}
Across Qwen checkpoints, Evidence accuracy without RegulaTutor rises from 20.0\% at $M_0$ to 28.2\% at $M_3$ (Table~\ref{tab:pairroute-main}), a paired gain of 8.2 points (95\% CI [6.3, 10.0]). Teaching rises from 2.631 to 2.789, a gain of 0.158 points (95\% CI [.088, .226]). Both endpoint scores are included in Table~\ref{tab:pairroute-main}.
Together with the monotonic development-set Evidence trajectory, these results show that reviewed execution traces provide useful supervision for evidence acquisition and instructional response quality beyond inference-time scaffolding.

\input{figures/table_pairroute_main}
\FloatBarrier

%% file: figures/table_longtutor_main.tex
\begin{table}[!t]
  \centering
  \caption{LongTutor Human Gold1000 (199-record histories). RegulaTutor uses final harness $H_1$. Green arrows show gains over model-only inference; mean rows average gains within each group. Evidence and Diagnosis gains are percentage points; Teaching gains are score points. Bold marks each model's best score; $^{*}$ marks a positive gain with a paired 95\% interval excluding zero. Codex uses GPT-5.6-sol; $^\dagger$ marks the backbone also used as Teaching judge.}
  \label{tab:longtutor-main}
  \small
  \definecolor{ltgain}{HTML}{237044}
  \newcommand{\ltup}[1]{\textcolor{ltgain}{\ensuremath{\,(\uparrow #1)}}}
  \setlength{\tabcolsep}{2.8pt}
  \begin{tabular*}{\linewidth}{@{\extracolsep{\fill}}llrrr@{}}
    \toprule
    \textbf{Model / system} & \textbf{Method} & \textbf{Evidence} & \textbf{Diagnosis} & \textbf{Teaching} \\
    \midrule
    \multicolumn{5}{@{}l}{\textit{Local Qwen checkpoints} ($n=4$)} \\
    \addlinespace[2pt]
    Qwen3.5-4B ($M_0$) & Model only & 20.0 & 29.7 & 2.63 \\
                       & RegulaTutor & \textbf{56.2}$^{*}$\ltup{36.2} & \textbf{50.8}$^{*}$\ltup{21.1} & \textbf{2.76}\ltup{0.13} \\
    \midrule
    Qwen3.5-4B ($M_1$) & Model only & 21.2 & 31.3 & 2.67 \\
                       & RegulaTutor & \textbf{56.4}$^{*}$\ltup{35.2} & \textbf{49.6}$^{*}$\ltup{18.3} & \textbf{2.73}\ltup{0.06} \\
    \midrule
    Qwen3.5-4B ($M_2$) & Model only & 27.9 & 28.0 & \textbf{2.79} \\
                       & RegulaTutor & \textbf{43.9}$^{*}$\ltup{16.0} & \textbf{51.6}$^{*}$\ltup{23.6} & 2.72 \\
    \midrule
    Qwen3.5-4B ($M_3$) & Model only & 28.2 & 27.2 & \textbf{2.79} \\
                       & RegulaTutor & \textbf{43.7}$^{*}$\ltup{15.5} & \textbf{51.3}$^{*}$\ltup{24.1} & 2.73 \\
    \cmidrule(lr){1-5}
    \multicolumn{2}{@{}l}{\textit{Mean RegulaTutor gain}} & \textcolor{ltgain}{+25.7} & \textcolor{ltgain}{+21.8} & \textcolor{ltgain}{+0.016} \\
    \midrule
    \multicolumn{5}{@{}l}{\textit{API backbones} ($n=5$)} \\
    \addlinespace[2pt]
    DeepSeek-v4-pro & Model only & 41.5 & 48.2 & 3.11 \\
                    & RegulaTutor & \textbf{52.3}$^{*}$\ltup{10.9} & \textbf{49.1}\ltup{0.9} & 3.33\ltup{0.22} \\
                    & DeepTutor & 48.7 & 37.9 & \textbf{3.89} \\
    \midrule
    Gemini-3.7-flash & Model only & 57.4 & \textbf{45.5} & 3.41 \\
                     & RegulaTutor & \textbf{75.6}$^{*}$\ltup{18.2} & 37.6 & \textbf{3.92}\ltup{0.51} \\
    \midrule
    GPT-5.6-sol & Model only & \textbf{54.9} & 37.7 & \textbf{3.56} \\
                & RegulaTutor & 50.1 & 41.0\ltup{3.3} & 3.26 \\
                & Native Codex & 44.2 & \textbf{42.8} & 3.43 \\
    \midrule
    Kimi-k3 & Model only & \textbf{47.2} & 25.8 & 3.05 \\
            & RegulaTutor & 46.0 & \textbf{33.5}$^{*}$\ltup{7.7} & \textbf{3.75}\ltup{0.70} \\
    \midrule
    DeepSeek-v4.1-flash$^\dagger$ & Model only & \textbf{56.4} & 29.3 & 3.03 \\
                                  & RegulaTutor & 48.9 & \textbf{42.2}$^{*}$\ltup{12.9} & \textbf{3.67}\ltup{0.64} \\
    \cmidrule(lr){1-5}
    \multicolumn{2}{@{}l}{\textit{Mean RegulaTutor gain}} & \textcolor{ltgain}{+3.1} & \textcolor{ltgain}{+3.4} & \textcolor{ltgain}{+0.353} \\
    \midrule
    \multicolumn{2}{@{}l}{\textit{Overall mean gain} ($n=9$)} & \textcolor{ltgain}{+13.2} & \textcolor{ltgain}{+11.6} & \textcolor{ltgain}{+0.203} \\
    \bottomrule
  \end{tabular*}
\end{table}

%% file: figures/table_harness_ablation.tex
\begin{table}[htbp]
  \definecolor{ablationred}{HTML}{B42318}
  \centering
  \caption{Component ablations in the completed LongTutor ablation batch. The within-batch reference scores are 37.6 Evidence, 44.9 Diagnosis, and 3.340 Teaching; absolute scores are not merged with the main $H_1$ run because the batch configuration differs. Parentheses report the ablated variant minus the reference.}
  \label{tab:harness-ablation}
  \normalsize
  \setlength{\tabcolsep}{4pt}
  \begin{tabular*}{\linewidth}{@{\extracolsep{\fill}}lrrr@{}}
    \toprule
    \textbf{Variant} & \textbf{Evidence} & \textbf{Diagnosis} & \textbf{Teaching} \\
    \midrule
    Full RegulaTutor & 37.6 & 44.9 & 3.340 \\
    Without CAMM interpretation & 35.6\,\textcolor{ablationred}{(-2.0)} & 44.8\,\textcolor{ablationred}{(-0.1)} & 3.176\,\textcolor{ablationred}{(-0.164)} \\
    Static policy & 36.1\,\textcolor{ablationred}{(-1.5)} & 44.8\,\textcolor{ablationred}{(-0.1)} & 3.196\,\textcolor{ablationred}{(-0.144)} \\
    Without compiled directives & 43.3\,(+5.7) & 44.2\,\textcolor{ablationred}{(-0.7)} & 3.228\,\textcolor{ablationred}{(-0.112)} \\
    \bottomrule
  \end{tabular*}
\end{table}

%% file: figures/table_pairroute_development.tex
\begin{table}[!htbp]
  \centering
  \caption{PairRoute trajectory on the repeatedly used dev100 set. $M_0H_0$ is the initial pair; the model-training series uses the selected harness, with $H_2=H_3=H_1$. Evidence is correct answers among 300 queries; Diagnosis is accuracy (\%); Teaching is the four-dimension mean. Bold indicates the highest value in each column.}
  \label{tab:pairroute-development}
  \normalsize
  \setlength{\tabcolsep}{5pt}
  \begin{tabular*}{\linewidth}{@{\extracolsep{\fill}}llrrrr@{}}
    \toprule
    \textbf{Pair} & \textbf{Model update} & \textbf{Evidence} & \textbf{Evidence (\%)} & \textbf{Diagnosis} & \textbf{Teaching} \\
    \midrule
    $M_0H_0$ & Initial model and harness & 208/300 & 69.3 & 43 & 2.690 \\
    $M_0H_1$ & Initial model, selected harness & 212/300 & 70.7 & 40 & 2.955 \\
    $M_1H_1$ & PairRoute SFT, round 1 & 230/300 & 76.7 & \textbf{50} & \textbf{3.055} \\
    $M_2H_1$ & PairRoute SFT, round 2 & 234/300 & 78.0 & \textbf{50} & 2.965 \\
    $M_3H_1$ & PairRoute SFT, round 3 & \textbf{250/300} & \textbf{83.3} & 48 & 2.935 \\
    \bottomrule
  \end{tabular*}
\end{table}

%% file: figures/table_pairroute_main.tex
\begin{table}[!htbp]
  \centering
  \caption{LongTutor Gold1000 results for Qwen3.5-4B checkpoints crossed with inference conditions. $M_0$ is the pretrained model; $M_1$--$M_3$ follow successive PairRoute stage-supervised training rounds. $H_0$ is the initial harness and $H_1$ the final RegulaTutor harness. E and D are Evidence and Diagnosis accuracy (\%); T is judged Teaching (1--5). Bold is reserved for the highest values supporting the positive results discussed in the main text.}
  \label{tab:pairroute-main}
  \normalsize
  \setlength{\tabcolsep}{3pt}
  \begin{tabular*}{\linewidth}{@{\extracolsep{\fill}}lrrrrrrrrr@{}}
    \toprule
    \textbf{Model checkpoint} & \multicolumn{3}{c}{\textbf{Model only}} & \multicolumn{3}{c}{\textbf{$H_0$}} & \multicolumn{3}{c}{\textbf{$H_1$}} \\
    \cmidrule(lr){2-4}\cmidrule(lr){5-7}\cmidrule(l){8-10}
     & \textbf{E} & \textbf{D} & \textbf{T} & \textbf{E} & \textbf{D} & \textbf{T} & \textbf{E} & \textbf{D} & \textbf{T} \\
    \midrule
    $M_0$ (pretrained) & 20.0 & 29.7 & 2.631 & 59.5 & 51.6 & 2.752 & 56.2 & 50.8 & \textbf{2.763} \\
    $M_1$ (round 1) & 21.2 & \textbf{31.3} & 2.671 & \textbf{67.3} & \textbf{52.7} & 2.721 & \textbf{56.4} & 49.6 & 2.731 \\
    $M_2$ (round 2) & 27.9 & 28.0 & \textbf{2.793} & 49.3 & 52.1 & 2.715 & 43.9 & \textbf{51.6} & 2.721 \\
    $M_3$ (round 3) & \textbf{28.2} & 27.2 & 2.789 & 59.9 & 52.0 & \textbf{2.774} & 43.7 & 51.3 & 2.733 \\
    \bottomrule
  \end{tabular*}
\end{table}

%% file: sections/related_work.tex
\section{Related Work}
\label{sec:related}
\paragraph{Intelligent Tutoring Systems}
Learning theory and learner modeling have long informed intelligent tutoring systems. MetaTutor supports metacognitive strategies through assessment and feedback \citep{metatutor2010}, while Wayang Outpost/MathSpring addresses cognition, metacognition, and affect in adaptive mathematics tutoring \citep{arroyo2014}. More recently, CLASS explicitly connects learning-science principles with the design of LLM-based tutoring \citep{class2023}. These systems provide foundations for connecting observations of learner activity to instructional support.

LLM tutoring frameworks extend these ideas with flexible generation, planning, and memory.
DeepTutor combines grounded tutoring with dynamic learner memory \citep{zhaoDeepTutorAgenticPersonalized2026}.
ScaffoldLM maintains a pedagogical plan and updates tutoring memory through assessment of learner progress \citep{scaffold2026}.
On the model side, \citet{tutorrl2025} train tutors with simulated interactions and rewards that balance pedagogical support and student problem solving.
Complementing these system and model developments, LongTutor evaluates the use of learning histories across evidence acquisition, diagnosis, and teaching \citep{liLongTutorBenchmarkingLarge2026}.
Together, these directions motivate examining how historical evidence informs teaching and how the model executes the resulting support.

Building on these foundations, RegulaTutor connects longitudinal evidence management with theory-informed instructional decisions.
Its harness combines multi-stage evidence resolution, CAMM-guided interpretation, persistent learner-specific teaching policies, and learner-facing SRL support.
Within this workflow, stored observations support learner interpretations, and policy guidelines translate those interpretations into instructional directives.
Incremental session memory allows learner history and policies to be maintained across interactions.

\paragraph{Model--Harness Co-Evolution}
A model update and a harness update modify different parts of an agent. The former changes reusable execution capabilities, whereas the latter changes the information and procedures available during execution. Their effects need not be independent. \citet{linHarnessUpdatingNot2026} distinguish the ability to produce useful harness updates from the ability to benefit from an updated harness, including failures to activate or follow relevant artifacts. This distinction motivates evaluating the executing model and its interface together.

HarnessForge represents an agent as a harness--policy pair and combines fault-guided harness tailoring with harness-conditioned policy alignment \citep{chenHarnessForgeJointHarness2026}. Co-Harness alternates validated local harness updates with model optimization on trajectories produced by the improved harness \citep{chenCoHarnessCoEvolvingHarnesses2026}. SIA similarly uses feedback to update both external scaffolding and model weights \citep{sia2026}. Failure-guided adaptation and joint optimization are direct precedents for our pipeline; they motivate allocating repairs according to the component responsible for a failure and validating the resulting pair.

Within RegulaTutor, PairRoute applies failure-guided offline refinement to historical evidence acquisition.
It diagnoses failures before harness revision and reassesses them under the selected harness, checking the input scope, output requirements, and evidence available at actual stage calls.
Confirmed residual model failures supply stage-level correction targets, which are combined with reviewed current-model outputs for retention.
This design connects component-level attribution to supervision under the selected harness, while monitoring downstream tutoring performance.

%% file: sections/discussion.tex
\section{Discussion}
\label{sec:discussion}

\paragraph{Revising learner interpretations over time.}
A past error remains a valid observation even after a learner has overcome the difficulty, and continuing to remediate it may then be inappropriate.
Long-term personalization therefore requires managing the lifetime of an interpretation separately from that of its supporting records.
RegulaTutor's evidence-linked policies make this distinction inspectable.
A useful extension is to associate each interpretation with both its supporting evidence and conditions for revision, so that accumulating history enables a tutor to revise outdated interpretations.

\paragraph{Evaluating evidence accuracy and teaching quality jointly.}
In the ablation batch, removing compiled directives increases Evidence while reducing judged Teaching (Table~\ref{tab:harness-ablation}).
This exposes a distinction between recovering historical facts and deciding how those facts should shape instruction.
A component's pedagogical value can therefore be missed by an evidence-only evaluation.
Future optimization could seek evidence-processing gains subject to explicit Diagnosis and Teaching criteria, keeping trade-offs visible when selecting model or harness updates.

\paragraph{Maintaining personalization across system updates.}
The crossed checkpoint results show that the effects of model updates depend on the inference configuration (Table~\ref{tab:pairroute-main}).
In a long-term tutoring service, learner histories and policies may persist after the model or harness changes.
This creates a continuity requirement: the updated system should still interpret and execute those policies appropriately.
Future compatibility tests could replay held-out learner histories together with their stored policies before and after an update, checking whether still-supported decisions are preserved in the resulting instruction.
The evaluation unit would then include persistent learner state alongside the model and harness.

%% file: sections/conclusion.tex
\section{Conclusion}
We presented \textit{RegulaTutor}, a theory-informed framework connecting learner history, evidence-supported interpretations, and adaptive teaching policies, together with \textit{PairRoute} for trace-guided offline refinement.
LongTutor evaluations show average gains in evidence acquisition and diagnosis, with Teaching gains on several backbones.
The refinement study shows development-set gains in evidence processing and characterizes model--harness compatibility.
Together, these contributions offer a foundation for exploring lifelong personalized tutoring in which instructional support can evolve alongside learners as their knowledge, goals, and learning contexts change.

\section*{Limitations}
LongTutor evaluates independent history snapshots and does not measure real learners' retention, transfer, or progress over time.
Teaching scores rely on model judges, with some configurations sharing the evaluator's model family.

The refinement study uses one local backbone, three supervised rounds, and an adaptively reused development set.
Crossed model--harness evaluations characterize compatibility, without budget-matched controls for alternative refinement strategies.
GRPO and stage-level SFT also differ in supervision, objectives, budgets, and checkpoint selection, preventing attribution of score differences to the optimization method alone.

Finally, the harness represents learning-related evidence through engineering heuristics; it does not provide calibrated estimates of psychological traits or establish causal effects of instructional interventions.

%% file: sections/statements.tex
\subsection*{AI use statement}
We used generative AI tools to polish the writing and improve readability. We reviewed all AI-assisted edits and take responsibility for the final content of this work.

\subsection*{Ethics statement}
Tutoring systems make inferences about learners that may be incomplete or wrong. Our design separates observed information from episode-specific interpretations and discourages unsupported claims about motivation or affect. These safeguards do not establish fairness or educational safety. Use with real learners requires appropriate data permissions, privacy protections, and review of the consequences of instructional decisions. This draft reports no new human-participant study and makes no claim of institutional approval. Any subsequent study must document its actual review and consent procedures.

\subsection*{Reproducibility statement}
Appendices~\ref{app:implementation}--\ref{app:benchmark_protocols} specify the current implementation boundary, two-stage round procedure, and benchmark protocols. A companion evidence ledger maps draft claims to local sources, server implementation records, and primary literature. The final release should include immutable model and harness revisions, learner-level split manifests, compiled snapshots, evaluator settings, and outputs for each pair comparison. Formal result admission requires matched generation/scoring artifacts and an overlap audit; obsolete or incompatible runs are excluded rather than used to fill missing results. No anonymous public artifact link is claimed in this version.

%% file: sections/appendix.tex
\definecolor{appblue}{HTML}{245875}
\definecolor{appgreen}{HTML}{34685B}
\definecolor{appgray}{HTML}{4A5563}
\definecolor{appamber}{HTML}{996515}
\definecolor{apppendingred}{HTML}{B42318}
\providecommand{\appnote}[1]{\par\smallskip\noindent{\color{appamber}\textbf{[To complete]} #1}\par\smallskip}
\providecommand{\appbenchmarkpending}[1]{\par\medskip\noindent{\color{apppendingred}\textbf{[BENCHMARK RESULTS TO ADD]} #1}\par\medskip}
\providecommand{\apprunpending}[1]{\par\smallskip\noindent{\color{appamber}\textbf{[FINAL RUN DETAILS TO VERIFY]} #1}\par\smallskip}
\providecommand{\appmissing}{\textcolor{appamber}{\textemdash}}
\providecommand{\appcode}[1]{\texttt{\detokenize{#1}}}

\section{RegulaTutor Implementation Details}
\label{app:implementation}

\subsection{History and Evidence Management}
\label{app:history}
RegulaTutor maintains a learner history $\mathcal{L}_{u,i}$ across interactions and selects evidence from it for the current task. Each learning event contains its content, timestamp, and source identifier, with concept and correctness fields when available.

We support two history modes. In \emph{snapshot mode}, each request supplies the complete history for an independent task, and learner state is reconstructed from that history without importing a stored policy. For example, a LongTutor case supplies past attempts and asks about a separate current question. In \emph{incremental mode}, a request supplies a new observation or dialogue turn, which is combined with the history and policy already stored for the same learner and session. This supports an ongoing tutoring conversation without resending the entire history. Events are deduplicated and ordered chronologically, and the updated state is saved after a successful interaction. Tutor-generated dialogue remains separate from scored learner observations, so an unscored exchange does not create a mastery label.

\paragraph{Evidence acquisition.}
\label{app:erl}
The Evidence Resolution Loop (ERL) obtains source evidence $E_{u,i}$ and answers $A_{u,i}$ to the task's information needs. A question about a particular attempt uses focused retrieval. A question spanning several sessions uses mapper calls to select relevant observations from history chunks, followed by consolidation and resolution. Each resolution includes supporting source identifiers. The harness checks these references and the output structure, and may expand retrieval or repeat resolution when support remains insufficient. This procedure bounds evidence acquisition while leaving interpretation and reasoning to the model.

\begin{table}[!htbp]
\caption{Core runtime settings. History and retrieval budgets count events; output limits count tokens per model call.}
\label{tab:app-runtime}
\centering
\begin{tabularx}{\linewidth}{@{}>{\raggedright\arraybackslash}Xl@{}}
\toprule
\textbf{Setting} & \textbf{Value}\\
\midrule
Bounded history / recent-history window & 10,000 / 10 events\\
Initial retrieval / ERL evidence top-$k$ & 8 / 48\\
Mapper chunk size & 40 events\\
Maximum ERL rounds & 2\\
Internal output limit / temperature & 12,000 tokens / 0\\
CAMM interpretation / evidence admission & Deterministic / ERL-cited events\\
Maximum policy proposals & 4\\
Policy acceptance threshold / switching cost & 0.60 / 0.15\\
\bottomrule
\end{tabularx}
\end{table}

\subsection{From Learner Evidence to Teaching Decisions}
\label{app:camm}
CAMM connects observed learner evidence to instructional support, drawing on SRL and MASRL \citep{zimmerman2002,efklides2011,molenaarMeasuringSelfregulatedLearning2023}. It represents cognitive and metacognitive interpretations separately, with motivational and affective signals sharing an implementation field. Each interpretation records its supporting observations and the limits of the inference. Table~\ref{tab:app-camm} summarizes how these interpretations guide teaching.

\begin{table}[!htbp]
\caption{CAMM links between learner evidence and instructional support. Rules apply only to admitted observations; absence of a signal does not establish a learner trait.}
\label{tab:app-camm}
\centering
\small
\begin{tabularx}{\linewidth}{@{}>{\raggedright\arraybackslash}p{.27\linewidth}>{\raggedright\arraybackslash}X>{\raggedright\arraybackslash}X@{}}
\toprule
\textbf{Observed evidence} & \textbf{Instructional support} & \textbf{Inference boundary}\\
\midrule
Repeated errors on a concept & Stepwise explanation and a focused check, grounded in a prior attempt & Local difficulty does not imply global inability or low motivation\\
Supported prior success & Concise explanation and retrieval practice & Prior accuracy does not establish general mastery\\
Explicit strategy, reflection, confidence, or self-check signal & Strategy comparison and self-checking & An observed action does not establish a stable metacognitive trait\\
Explicit motivational or affective self-report & Acknowledge the signal and adapt support to its context & Incorrect answers alone do not establish motivation or emotion\\
\bottomrule
\end{tabularx}
\end{table}

\paragraph{Interpretation and policy updates.}
CAMM uses ERL-cited events and recomputes learner statistics over this admitted evidence. Repeated difficulty requires at least two incorrect outcomes on the same concept; prior-success support requires at least two scored events and accuracy of at least 0.7. Confidence values and acceptance thresholds are fixed engineering heuristics, rather than calibrated psychological probabilities.

RegulaTutor converts supported interpretations into policy proposals, ranked by confidence, guideline relevance, and evidence support. A switching cost favors continuity with the previous policy. When proposals conflict, the highest-ranked eligible assignment determines each policy field; the no-inference guard remains active even when other proposals are rejected. The resulting policy retains links to its supporting observations.

\paragraph{Instruction compilation.}
\label{app:runtime-procedure}
RegulaTutor combines the CAMM interpretation $C_{u,i}$ and policy $P_{u,i}$ into teaching directives $D_{u,i}$ specifying support mode, explanation depth, history references, and inferences to avoid. The final model input combines these directives with the current task, learner statistics, resolved evidence, and recent dialogue. A shared system prompt also encourages goal setting, self-checking, and reflection when appropriate. Learner history and policy can change between interactions, while model weights and shared harness rules remain fixed during tutoring.

\section{PairRoute Refinement Study and Model Training Details}
\label{app:algorithm}

\subsection{Failure Diagnosis and Harness Refinement}
\label{app:routing}
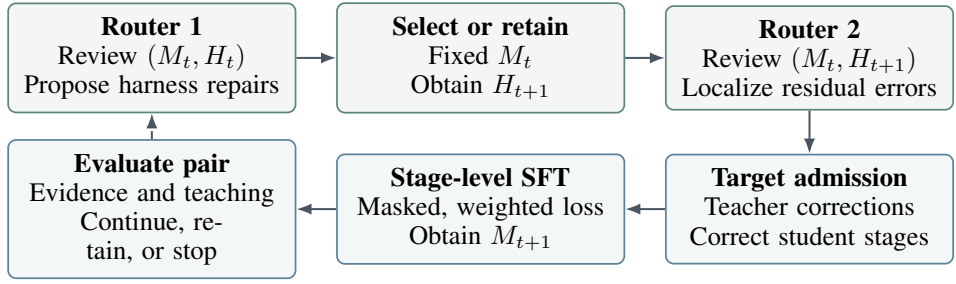
\begin{figure}[!htbp]
\centering
\begin{tikzpicture}[
box/.style={draw=appgreen!80,fill=appgreen!6,rounded corners=3pt,align=center,text width=3.45cm,minimum height=1.25cm,inner sep=5pt,line width=.65pt},
model/.style={box,draw=appblue!75,fill=appblue!5},
flow/.style={-{Latex[length=2mm]},line width=.75pt,draw=appgray}]
\node[box] (r1) at (0,0) {\textbf{Router 1}\\Review $(M_t,H_t)$\\Propose harness repairs};
\node[box] (h) at (4.35,0) {\textbf{Select or retain}\\Fixed $M_t$\\Obtain $H_{t+1}$};
\node[box] (r2) at (8.7,0) {\textbf{Router 2}\\Review $(M_t,H_{t+1})$\\Localize residual errors};
\node[model] (data) at (8.7,-2) {\textbf{Target admission}\\Teacher corrections\\Correct student stages};
\node[model] (train) at (4.35,-2) {\textbf{Stage-level SFT}\\Masked, weighted loss\\Obtain $M_{t+1}$};
\node[model] (eval) at (0,-2) {\textbf{Evaluate pair}\\Evidence and teaching\\Continue, retain, or stop};
\draw[flow] (r1)--(h); \draw[flow] (h)--(r2);
\draw[flow] (r2)--(data); \draw[flow] (data)--(train); \draw[flow] (train)--(eval);
\draw[flow,dashed] (eval)--(r1);
\end{tikzpicture}
\caption{A PairRoute round. Harness selection precedes the construction of model supervision. Trajectories may be reused when the selected model--harness pair is unchanged.}
\label{fig:app-pairroute}
\end{figure}

PairRoute studies evidence-processing refinement through candidate harness revision and model training. At round $t$, Router1 examines execution under $(M_t,H_t)$ and proposes repairs to evidence access or presentation. Candidates are evaluated with $M_t$ fixed to select or retain $H_{t+1}$. Router2 then reviews residual failures under $(M_t,H_{t+1})$ and identifies model calls eligible for supervision. Figure~\ref{fig:app-pairroute} summarizes the executed procedure. The current experiments evaluate the resulting trajectory and crossed model--harness pairs; they do not estimate an advantage over alternative routing or update algorithms.

\begin{table}[!htbp]
\caption{Routing decisions based on the evidence visible to the implicated model call.}
\label{tab:app-routes}
\centering
\begin{tabularx}{\linewidth}{@{}l>{\raggedright\arraybackslash}X@{}}
\toprule
\textbf{Disposition} & \textbf{Criterion and action}\\
\midrule
Harness & Evidence access or the call interface is defective; propose a harness repair.\\
Model & Evidence and instructions are sufficient, but execution is incorrect; review a correction target.\\
Mixed & Both defects occur; repair the interface before reassessing model supervision.\\
Success & The task meets its evaluation criterion; independently review stages for retention.\\
External & A source, reference, or evaluation problem prevents a clean correction target.\\
Defer & Attribution remains uncertain; withhold the case from confirmed corrections.\\
\bottomrule
\end{tabularx}
\end{table}

We instantiate the two routers as Codex agents with separate review instructions. They inspect the task, reference evaluation, and actual model-call inputs and outputs. The offline pipeline checks call identities, source references, and review structure; the agents judge semantic correctness. Router1's proposed repairs receive manual refinement before candidate evaluation. Router2 reviews calls under the selected harness so that training targets reflect the interface used by the student.

\subsection{Stage-Level Supervision and Training}
\label{app:training}
We construct training data from the 369-case discovery subset of the LongTutor XES3G5M silver-data pipeline, separate from Human Gold1000 used for benchmark evaluation (Section~\ref{app:splits}). Each discovery case supplies 199 history records, a current question, and three evidence queries: information extraction (IE), multi-session reasoning (MR), and hallucination control (HC). We execute the student under the selected harness on these cases and use the resulting calls to construct supervision; the silver reference answers are evidence for review rather than targets copied directly into training.

Answering a query may involve several model calls. A \emph{stage-level} example supervises one call---a mapper selecting records, a resolver answering from evidence, an audit checking support, or a final response generator---using its recorded input and a reviewed target.

The training set $\mathcal D_t=\mathcal D_t^{\mathrm{corr}}\cup\mathcal D_t^{\mathrm{ret}}$ combines teacher corrections with correct outputs from the current student. Retention is reviewed at the call level, including correct stages within failed tasks. Teacher targets must be valid in full, even when only part of the output receives loss. If a teacher trajectory uses different upstream context, its compatibility with the training obligation is checked explicitly. Hidden references and routing judgments are excluded from student inputs.

\paragraph{Teacher and loss.}
Kimi K3 supplies teacher targets with low thinking effort, temperature~1, and a 12,000-token output limit. Compatible reviewed targets may be reused across rounds. For input $x_r$, target $y_r$, admitted-token mask $m_{r\ell}$, and token weight $w_{r\ell}$, the microbatch objective is
\begin{equation}
\mathcal L_{\mathcal B}(\theta)=
-\frac{\sum_{r\in\mathcal B}\sum_{\ell}m_{r\ell}w_{r\ell}
\log p_\theta(y_{r\ell}\mid x_r,y_{r,<\ell})}
{\sum_{r\in\mathcal B}\sum_{\ell}m_{r\ell}w_{r\ell}}.
\label{eq:app-stage-loss}
\end{equation}
Prompt and padding tokens receive no loss. We balance total supervised-token mass across IE, MR, and HC; the three categories need not have equal numbers of examples.

\paragraph{Training configuration.}
All updates use LoRA with rank~16, scaling factor~32, and dropout~0.05 on attention, linear-attention, and feed-forward projections. We use AdamW, a linear learning-rate schedule with 5\% warmup, gradient clipping at~1, BF16 precision, and gradient checkpointing. Each update runs for two epochs with global batch size~4, seed~42, and a maximum sequence length of 34,000 tokens; no admitted example is truncated. The first update initializes a new adapter on Qwen3.5-4B. Later updates continue the preceding adapter with a fresh optimizer. Every round uses the final epoch, without development-based checkpoint selection.

\begin{table}[!htbp]
\caption{Training data and model updates. Corrections and retention count unique stage examples; each corpus has equal IE/MR/HC supervised-token mass.}
\label{tab:app-training-data}
\label{tab:app-training-config}
\label{tab:app-rounds}
\centering
\begin{tabular*}{0.90\linewidth}{@{\extracolsep{\fill}}lrrrrl@{}}
\toprule
\textbf{Update} & \textbf{Cases} & \textbf{Corrections} & \textbf{Retention} & \textbf{Total} & \textbf{Learning rate}\\
\midrule
$M_0\rightarrow M_1$ & 256 & 264 & 137 & 401 & $2\times10^{-5}$\\
$M_1\rightarrow M_2$ & 121 & 87 & 70 & 157 & $10^{-5}$\\
$M_2\rightarrow M_3$ & 111 & 60 & 60 & 120 & $10^{-5}$\\
\bottomrule
\end{tabular*}
\end{table}

\subsection{Selected Harness and Refinement Rounds}
\label{app:versions}
$M_0$ denotes Qwen3.5-4B and $H_0$ the initial harness with incremental learner-state management. The reported $H_0$ is frozen at tag \appcode{H\_0}, commit \appcode{81a7e37}, with configuration-hash prefix \appcode{88c2ad8f}. The first round produces $H_1$ through Router1-guided revision and manual refinement. Later harness candidates are not retained, so $H_1$ is the final PairRoute-refined harness reported as RegulaTutor in the main fixed-backbone results. It adds current-context binding, which clarifies the current task's referent, and an indexed factual-support audit. The audit checks answered resolutions for queries that a language-based detector does not classify as collection operations; detected collection queries bypass it. Both classification and support checking remain fallible model/harness operations.

The revisions target two evidence failures observed during development: binding a query to the wrong current context and accepting an answer without sufficient source support. After retaining $H_1$ for the next stage, PairRoute continues to review residual failures and construct model supervision. Later harness candidates do not meet the screening criteria, so $H_2=H_3=H_1$ while the model advances to $M_2$ and $M_3$. These version transitions describe the executed refinement trajectory; whether any candidate improves the complete system must be determined from the matched evaluations rather than inferred from its position in the sequence.

\subsection{Matched Comparisons}
\label{app:pair-controls}
To distinguish harness and model changes, adjacent-round comparisons consider $(M_{t-1},H_{t-1})$, $(M_{t-1},H_t)$, $(M_t,H_{t-1})$, and $(M_t,H_t)$ under the same evaluation protocol. For a score $s$, their interaction contrast is
\begin{equation}
\begin{split}
\Delta_{\mathrm{int},t}={}&s(M_t,H_t)-s(M_t,H_{t-1})\\
&-s(M_{t-1},H_t)+s(M_{t-1},H_{t-1}).
\end{split}
\label{eq:interaction}
\end{equation}
When the harness is retained, identical configurations collapse and the contrast is zero if the same observations are reused. In round~3, Router2 reuses the $(M_2,H_2)$ trajectories after verifying that they also represent $(M_2,H_3)$. These comparisons characterize the updates made within PairRoute; establishing a benefit from routing itself would require controls with matched data and review budgets.
Table~\ref{tab:pairroute-main} reports the completed full-benchmark checkpoint--harness grid. Because these are crossed evaluations of one training trajectory rather than independently optimized control branches, we use them to characterize compatibility and do not estimate $\Delta_{\mathrm{int},t}$ as evidence that routing outperforms an alternative update algorithm.

\section{Datasets and Evaluation Protocols}
\label{app:benchmark_protocols}
\label{app:benchmark}

This appendix records the supplementary protocols and analyses that are not used as primary claims in the main text. LongTutor is the primary benchmark for the paper. TeachArena is retained as an exploratory transfer study because it exercises simulated interaction and tool execution under a different runtime; its results should be read as protocol-specific evidence about interface compatibility, rather than as a second main benchmark.

\subsection{Benchmarks and Metrics}
\paragraph{LongTutor.}
We use XES3G5M Human Gold1000, comprising 1,000 cases with 199 history records and a separate current question per case \citep{liLongTutorBenchmarkingLarge2026}. All systems receive the same public history and task inputs. Reference answers, hidden diagnostic statistics, and evaluator metadata are withheld. IE, MR, and HC are scored separately, and their arithmetic mean is the Evidence score. Diagnosis uses accuracy and macro-F1 over four classes. Teaching uses the mean of History Utilization, Strategy Alignment, Coherence, and Appropriateness, each scored on a 1--5 scale. The separately returned overall judgment is an auxiliary measure.

\paragraph{TeachArena.}
We run the \texttt{edu\_agent} domain in the TeachArena $\tau^2$-bench fork with \texttt{EDU\_CANVAS\_MODE=mock} \citep{chenTeachArenaAreLanguage2026}. We report Paper Stage~2 (repository split \texttt{s0}), which contains 114 situated multi-turn tutoring tasks, and Paper Stage~3 (repository split \texttt{s2}), which contains 136 Canvas workflow tasks. Paper Stage~1 was not executed and is not included in any aggregate. Stage~2 evaluates the dialogue trajectory with a simulated learner; Stage~3 evaluates tool-mediated actions and the resulting LMS state. Rewards lie in $[0,1]$.

\subsection{Data Splits}
\label{app:splits}
PairRoute uses separate discovery, probe, and development sets from the XES3G5M silver-data pipeline (Table~\ref{tab:app-splits}). For harness refinement, Router1 diagnoses failures on discovery trajectories and proposes interface changes. Candidates are tested with the model fixed, using selected discovery cases for targeted checks and probe for screening. Development assesses the selected model--harness pairs across rounds. For model training, Router2 reviews discovery trajectories under the selected harness and constructs stage supervision. The probe and development sets are reused adaptively during the executed refinement procedure.

The full-benchmark harness comparison evaluates direct inference, the initial harness $H_0$, and the refined harness $H_1$ with a fixed backbone. Component ablations assess the contributions of evidence resolution, CAMM, policy adaptation, and compiled directives (Section~\ref{app:ablations}). These benchmark comparisons are distinct from the candidate-screening experiments used to select a harness.

\begin{table}[!htbp]
\caption{Data used for refinement and evaluation. The three LongTutor adaptation sets are pairwise disjoint and have no overlap with Human Gold1000 in case, learner, or complete-history identity.}
\label{tab:app-splits}
\centering
\begin{tabularx}{\linewidth}{@{}l r >{\raggedright\arraybackslash}X@{}}
\toprule
\textbf{Set} & \textbf{Cases/tasks} & \textbf{Purpose}\\
\midrule
Discovery & 369 & Harness failure diagnosis, targeted candidate checks, and stage supervision\\
Probe & 32 & Harness candidate screening\\
Development & 100 & Adaptive assessment of model--harness updates\\
LongTutor Gold1000 & 1,000 & System, harness, and model--harness pair evaluation\\
TeachArena dialogue & 114 & Situated multi-turn tutoring\\
TeachArena workflows & 136 & Tool-mediated Canvas tasks\\
\bottomrule
\end{tabularx}
\end{table}

We verify separation using case and learner identifiers and hashes of complete histories. Each LongTutor set contains one case per learner, and all admitted training episodes belong to discovery. TeachArena contributes no examples to the three stage-SFT corpora. These checks concern the refinement data, not the backbone's pretraining exposure.

\subsection{Evaluation Settings}
LongTutor cases use independent history snapshots. Its judge is \appcode{deepseek-v4.1-flash}, run at temperature zero. Invalid outputs remain in the 1,000-case denominator; the Kimi-k3/$H_1$ arm has 126 such cases. For paired comparisons, we align arms by case identifier and bootstrap the case-level difference with 3,000 resamples (seed 20260921), reporting percentile 95\% intervals. Evidence, Diagnosis, and Teaching each use all 1,000 paired cases; Teaching is included only after all four rubric scores have been completed for every record. The intervals are descriptive and receive no multiplicity correction. PairRoute development comparisons instead use Codex reviews under fixed rubrics and are kept separate from the full benchmark.

The original content-scoring pass left 9,846 judgments missing across 20 arms after evaluator calls reached the routing service's rate limit. We re-issued only these missing calls with the same judge and decoding settings, retained every existing judgment unchanged, and merged the repaired records by case identifier. The completed records contain all four Teaching rubrics for every case; the original and repaired files remain separately archived.

TeachArena retains dialogue and public tool feedback within a task and resets state between tasks. The student simulator is \appcode{qwen3-max}; the judge is \appcode{gpt-5.1-2025-11-13} at temperature zero. Each agent receives at most 60 steps. Direct conditions use the stock \texttt{llm\_agent}; $H_1$ wraps the same backbone as a $\tau^2$ agent. We report both full-set means, where failed executions receive zero, and paired means over task identifiers completed by both conditions. Paired 95\% confidence intervals resample task identifiers jointly across the two conditions and describe sample uncertainty for the recorded runs, rather than variation across seeds or judges.

\section{Baselines and Ablation Protocol}
\label{app:baselines}

\subsection{Baseline Configurations}
We compare model-only inference, DeepTutor, native Codex, the initial harness $H_0$, and the final PairRoute-refined harness $H_1$. Model-only inference exposes the model to the public task inputs without RegulaTutor's intermediate stages. DeepTutor supplies a tutoring-harness baseline on DeepSeek-v4-pro; native Codex supplies a GPT-5.6-sol agent configuration with its own runtime. The main RegulaTutor results use $H_1$; the appendix retains $H_0/H_1$ comparisons to characterize the harness revision and model--harness compatibility. Table~\ref{tab:app-baselines} summarizes the current evaluation coverage.

\begin{table}[!htbp]
\caption{System configurations and recorded evaluation coverage. LT denotes LongTutor; TA-D and TA-W denote TeachArena dialogue and workflow tasks.}
\label{tab:app-baselines}
\centering
\small
\begin{tabularx}{\linewidth}{@{}l>{\raggedright\arraybackslash}X>{\raggedright\arraybackslash}X@{}}
\toprule
\textbf{System} & \textbf{Recorded coverage} & \textbf{Configuration boundary}\\
\midrule
Direct inference & LT: nine model/checkpoint settings; TA-D/TA-W: three backbones & Public inputs and benchmark agent contract\\
DeepTutor & LT: deepseek-v4-pro only & Real v1.5.1 chat engine; task-local sessions\\
Native Codex & LT: GPT-5.6-sol agent condition & Own agent runtime and execution budget\\
RegulaTutor $H_0$ & LT: $M_0$--$M_3$ and deepseek-v4-pro & Frozen initial harness, distinct from direct inference\\
RegulaTutor $H_1$ & LT: nine settings; TA-D/TA-W: three backbones & Final PairRoute-refined harness; later $H_2,H_3$ retain it\\
\bottomrule
\end{tabularx}
\end{table}

LongTutor uses the public 199-record histories and scoring protocol in Section~\ref{app:benchmark_protocols}. The condition using the tested model as Teaching judge is identified in result tables and included in descriptive group means. Although native Codex uses GPT-5.6-sol, its distinct runtime means that comparison with model-only inference or RegulaTutor does not isolate a harness effect.

GPT and Gemini use JSON-object generation after strict JSON-schema requests are rejected. Response validation and retry settings are therefore recorded alongside decoding parameters. The prepared protocol permits 3,000 output tokens for direct and harness final responses and 12,000 for internal harness calls. DeepTutor and Codex use their respective execution budgets. We account for all calls and retries rather than assuming equal computation.

\subsection{Ablation Definitions}
\label{app:ablations}
Each ablation uses the same backbone, task set, evaluator, and starting state within the ablation batch. Table~\ref{tab:app-ablations} defines the interventions, and Table~\ref{tab:harness-ablation} reports the corresponding aggregate contrasts. The ablation batch has a different absolute reference score from the main $H_1$ run, so we interpret its paired differences rather than merge its absolute values with the main table.

\begin{table}[!htbp]
\caption{Component interventions for testing the evidence-to-instruction pathway. The completed score table is reported in Table~\ref{tab:harness-ablation}.}
\label{tab:app-ablations}
\centering
\small
\begin{tabularx}{\linewidth}{@{}>{\raggedright\arraybackslash}p{.19\linewidth}>{\raggedright\arraybackslash}X>{\raggedright\arraybackslash}X@{}}
\toprule
\textbf{Variant} & \textbf{Intervention} & \textbf{What the contrast measures}\\
\midrule
Without CAMM pathway & Disable interpretation and its downstream policy/directive pathway; retain evidence inputs and static support & Combined contribution of the CAMM-conditioned pathway, not the compiler alone\\
Static policy & Keep CAMM and its direct implications but fix policy fields to their default values & Contribution of adaptive policy-field selection beyond direct CAMM implications\\
Without compiled directives & Keep internal interpretation/policy computation but remove compiled directives from the final prompt; freeze other field exposure & Contribution of explicit instruction compilation under the recorded exposure policy\\
\bottomrule
\end{tabularx}
\end{table}
The completed batch shows positive support for CAMM interpretation and adaptive policy selection: removing CAMM reduces Evidence by 2.03 points and Teaching by 0.164, while a static policy reduces Evidence by 1.50 points and Teaching by 0.144. Diagnosis changes are not detected for these contrasts.

\subsection{GRPO Training-Path Probe}
\label{app:grpo}
As a supplementary training-path probe, we compare two cumulative GRPO checkpoints from a separate training batch under the same recorded $H_1$ interface. G1 is the first GRPO update and G2 is a second adaptive update from G1; both use LoRA adapters over the same frozen base. Within this batch, Evidence increases from 47.23\% to 64.57\%, a paired gain of 17.33 points (95\% CI [15.9, 18.7]), while Diagnosis and Teaching remain statistically unchanged. This result shows that the training path can interact strongly with the evidence interface; its absolute scores are not merged with the main SFT checkpoint table.
\input{figures/table_grpo_probe}
\FloatBarrier

\subsection{Interface Controls}

For TeachArena workflows, output validity and task reward measure different aspects of performance. Table~\ref{tab:app-interface-control} reports response-contract variants recorded within the $H_1$ agents. Relaxing strict schema enforcement sharply improves completion for DeepSeek and Gemini, showing that interface reliability must be separated from instructional quality.

\begin{table}[!htbp]
\caption{TeachArena workflow response-mode controls within $H_1$. Reward uses all 136 tasks; scored cases reached evaluation. The dash denotes an unavailable count, not zero.}
\label{tab:app-interface-control}
\centering
\small
\begin{tabular*}{0.85\linewidth}{@{\extracolsep{\fill}}llrr@{}}
\toprule
\textbf{Backbone} & \textbf{Response mode} & \textbf{Reward} & \textbf{Scored cases}\\
\midrule
Qwen3.5-4B & Strict schema & .282 & 130/136\\
             & JSON object & .001 & 2/136\\
DeepSeek-v4-pro & Strict schema & .430 & 101/136\\
                & JSON object & .531 & 127/136\\
Gemini-3.7-flash & Strict schema & $\approx.000$ & \appmissing\\
                  & JSON object & .554 & 120/136\\
\bottomrule
\end{tabular*}
\end{table}

The response-mode effect differs across backbones. Qwen's JSON-object run records 93 unterminated-string errors and only 2/136 scored tasks. For DeepSeek, relaxing the schema reduces unscored executions from 35 to 9. Gemini's strict-schema condition produces empty tool arguments; its exact scored-task count is not available in the report. These controls change the response mode within $H_1$ and do not constitute a matched Direct/$H_1$ validation-and-retry ablation. The main transfer comparison uses the better-performing reported mode for each backbone, so it characterizes the selected interface configuration.

\FloatBarrier
\section{Additional Results and Analyses}
\label{app:additional-results}
This section provides analyses that complement the main results without repeating their aggregate tables. All paired intervals use the same 1,000 LongTutor cases and the completed Teaching judgments described in Section~\ref{app:benchmark_protocols}. Development scores use a different split and evaluator and are reported only in Table~\ref{tab:pairroute-development}.

\subsection{LongTutor: Paired and Query-Level Effects}
\label{app:weekly-summary}
Table~\ref{tab:longtutor-main} gives the aggregate Evidence, Diagnosis, and Teaching scores. Table~\ref{tab:app-longtutor-paired} adds paired uncertainty for each fixed-backbone contrast; Table~\ref{tab:app-evidence-query-paired} separates the three Evidence query types. Evidence averages information extraction (IE), multi-session reasoning (MR), and hallucination control (HC); Diagnosis is four-class accuracy; Teaching averages History Utilization, Strategy Alignment, Coherence, and Appropriateness on a 1--5 scale. Invalid generations remain in the 1,000-case denominator for Evidence and Diagnosis.

\begingroup
\setlength{\intextsep}{8pt}
\setlength{\abovecaptionskip}{4pt}
\setlength{\belowcaptionskip}{4pt}

\begin{table}[H]
\caption{Paired effects of the final RegulaTutor harness $H_1$ relative to Direct on LongTutor. Evidence and Diagnosis differences are percentage points; Teaching differences use the 1--5 scale. Every interval uses 3,000 paired case-bootstrap resamples over the same 1,000 cases. $^\dagger$ marks the evaluator backbone, excluded from across-backbone averages.}
\label{tab:app-longtutor-paired}
\centering
\small
\setlength{\tabcolsep}{2.6pt}
\begin{tabular*}{\linewidth}{@{\extracolsep{\fill}}lrrr@{}}
\toprule
\textbf{Backbone} & \textbf{Evidence $\Delta$ [95\% CI]} & \textbf{Diagnosis $\Delta$ [95\% CI]} & \textbf{Teaching $\Delta$ [95\% CI]}\\
\midrule
Qwen $M_0$ & +36.17 [34.1, 38.2] & +21.10 [17.2, 24.9] & +.132 [.064, .200]\\
Qwen $M_1$ & +35.23 [33.1, 37.2] & +18.30 [14.1, 22.3] & +.060 [--.006, .127]\\
Qwen $M_2$ & +16.00 [13.9, 18.1] & +23.60 [19.3, 27.9] & --.072 [--.138, --.006]\\
Qwen $M_3$ & +15.50 [13.4, 17.6] & +24.10 [19.9, 28.2] & --.057 [--.124, .012]\\
DeepSeek-v4-pro & +10.87 [8.7, 13.0] & +.90 [--2.1, 3.8] & +.220 [.172, .268]\\
Gemini-3.7-flash & +18.23 [16.6, 20.0] & --7.90 [--11.7, --4.1] & +.506 [.458, .553]\\
GPT-5.6-sol & --4.87 [--7.0, --2.8] & +3.30 [--.1, 6.9] & --.302 [--.347, --.255]\\
Kimi-k3 & --1.23 [--3.8, 1.3] & +7.70 [4.4, 10.9] & +.698 [.600, .791]\\
\midrule
DeepSeek-v4.1-flash$^\dagger$ & --7.50 [--9.4, --5.5] & +12.90 [9.5, 16.2] & +.645 [.591, .699]\\
\bottomrule
\end{tabular*}
\end{table}
\vspace{0pt}

\begin{table}[H]
\caption{Per-query Evidence differences for final RegulaTutor $H_1$ relative to Direct. Values are percentage points with 95\% paired bootstrap intervals over 1,000 cases. IE, MR, and HC denote information extraction, multi-session reasoning, and hallucination control. $^\dagger$ marks the evaluator backbone.}
\label{tab:app-evidence-query-paired}
\centering
\small
\setlength{\tabcolsep}{3pt}
\begin{tabular*}{\linewidth}{@{\extracolsep{\fill}}lrrr@{}}
\toprule
\textbf{Backbone} & \textbf{IE $\Delta$ [95\% CI]} & \textbf{MR $\Delta$ [95\% CI]} & \textbf{HC $\Delta$ [95\% CI]}\\
\midrule
Qwen $M_0$ & +43.2 [39.4, 47.0] & +17.9 [14.9, 20.8] & +47.4 [44.1, 50.7]\\
Qwen $M_1$ & +39.7 [35.8, 43.6] & +18.3 [15.3, 21.4] & +47.7 [44.6, 51.0]\\
Qwen $M_2$ & +2.5 [--1.5, 6.5] & +1.1 [--1.6, 3.8] & +44.4 [41.1, 47.7]\\
Qwen $M_3$ & +3.1 [--.8, 7.2] & +1.9 [--1.0, 4.7] & +41.5 [38.1, 44.7]\\
DeepSeek-v4-pro & --33.4 [--37.5, --29.2] & +23.6 [19.8, 27.2] & +42.4 [39.3, 45.5]\\
Gemini-3.7-flash & +25.0 [22.1, 27.9] & +25.3 [22.0, 28.5] & +4.4 [2.2, 6.6]\\
GPT-5.6-sol & --57.9 [--61.2, --54.7] & +9.5 [5.9, 13.1] & +33.8 [30.8, 36.7]\\
Kimi-k3 & --7.7 [--11.8, --3.7] & +5.6 [1.8, 9.3] & --1.6 [--4.8, 1.7]\\
\midrule
DeepSeek-v4.1-flash$^\dagger$ & --64.9 [--68.1, --61.7] & +18.7 [14.7, 22.4] & +23.7 [20.8, 26.6]\\
\bottomrule
\end{tabular*}
\end{table}

The query-level results qualify the aggregate Evidence gains: $M_0$, $M_1$, and Gemini improve on all three query types, while the response is uneven on other backbones. Hallucination control improves reliably on seven of the eight configurations judged independently of the tested backbone. Excluding the self-judged DeepSeek-v4.1-flash condition, the eight Direct/$H_1$ pairs average gains of 15.7 Evidence and 11.4 Diagnosis percentage points. These are descriptive averages over configurations, not an estimate of learner-level effects.

\endgroup

\subsection{Interactive Transfer and Interface Stress Test}
\label{app:interactive-transfer}
TeachArena tests simulated dialogue and tool-mediated workflows under a different protocol from LongTutor. Table~\ref{tab:app-teachar-results} reports full-task rewards and outcomes conditional on both configurations completing a task. The interface response-mode controls in Table~\ref{tab:app-interface-control} help interpret differences in execution coverage.

\begin{table}[H]
\caption{TeachArena results. Full means assign zero to failed executions. Paired means and confidence intervals use task identifiers completed by both Direct and $H_1$. W/L/T gives $H_1$ wins, losses, and ties.}
\label{tab:app-teachar-results}
\centering
\small
\setlength{\tabcolsep}{2.4pt}
\begin{tabular*}{\linewidth}{@{\extracolsep{\fill}}llrrrrrr@{}}
\toprule
\textbf{Task} & \textbf{Backbone} & \textbf{Direct} & \textbf{$H_1$} & \textbf{$n$} & \textbf{$\Delta$} & \textbf{95\% CI} & \textbf{W/L/T}\\
\midrule
Dialogue & Qwen3.5-4B & .572 & .587 & 106 & +.015 & [--.003,.034] & 36/24/46\\
 & DeepSeek-v4-pro & .672 & .621 & 107 & --.010 & [--.024,.002] & 22/29/56\\
 & Gemini-3.7-flash & .666 & .641 & 111 & --.007 & [--.021,.006] & 23/28/60\\
\midrule
Workflows & Qwen3.5-4B & .547 & .282 & 122 & --.290 & [--.362,--.217] & 18/87/17\\
 & DeepSeek-v4-pro & .682 & .531 & 127 & --.129 & [--.201,--.061] & 30/59/38\\
 & Gemini-3.7-flash & .646 & .554 & 120 & --.011 & [--.072,.051] & 29/43/48\\
\bottomrule
\end{tabular*}
\end{table}

Full-task rewards include failed executions as zero; paired intervals condition on tasks completed by both arms. Under $H_1$, dialogue failures are 4/114, 7/114, and 3/114 for Qwen, DeepSeek, and Gemini, respectively; workflow failures are 6/136, 9/136, and 16/136. All three dialogue paired intervals include zero, while Qwen and DeepSeek workflow intervals exclude zero in the negative direction. The paired intervals therefore cannot be read as uncertainty intervals for the full-task differences.

\subsection{Model--Harness Compatibility}
\label{app:benchmark-pairs}
The crossed Gold1000 scores appear in Table~\ref{tab:pairroute-main}; Table~\ref{tab:app-harness-compatibility} adds paired intervals for the $H_1-H_0$ contrast at fixed checkpoints. Direct inference is a separate public-input baseline, not $H_0$. Later harness aliases collapse to $H_1$ because $H_2=H_3=H_1$.

\begin{table}[H]
\caption{Harness compatibility on the same 1,000 cases. Differences are $H_1-H_0$; Evidence and Diagnosis use percentage points. Teaching uses score units. Intervals are unadjusted paired case-bootstrap intervals.}
\label{tab:app-harness-compatibility}
\centering\small
\setlength{\tabcolsep}{3pt}
\begin{tabular*}{\linewidth}{@{\extracolsep{\fill}}lrrr@{}}
\toprule
\textbf{Comparison} & \textbf{Evidence $\Delta$ [95\% CI]} & \textbf{Diagnosis $\Delta$ [95\% CI]} & \textbf{Teaching $\Delta$ [95\% CI]}\\
\midrule
$M_0$: $H_1-H_0$ & $-3.37\;[-5.30, -1.50]$ & $-0.80\;[-2.90, 1.20]$ & $+0.011\;[-0.035, 0.061]$\\
$M_1$: $H_1-H_0$ & $-10.83\;[-12.60, -9.10]$ & $-3.10\;[-5.10, -1.20]$ & $+0.010\;[-0.041, 0.061]$\\
$M_2$: $H_1-H_0$ & $-5.40\;[-7.30, -3.40]$ & $-0.50\;[-2.30, 1.30]$ & $+0.006\;[-0.045, 0.061]$\\
$M_3$: $H_1-H_0$ & $-16.20\;[-18.20, -14.30]$ & $-0.70\;[-2.40, 0.90]$ & $-0.041\;[-0.094, 0.011]$\\
DeepSeek: $H_1-H_0$ & $+1.97\;[-0.10, 4.10]$ & $+4.60\;[1.60, 7.70]$ & $-0.010\;[-0.067, 0.050]$\\
\bottomrule
\end{tabular*}
\end{table}

$H_1$ scores below $H_0$ on Evidence at every Qwen checkpoint, with all four paired intervals excluding zero; the largest decrease is 16.20 percentage points at $M_3$. Local Teaching contrasts are inconclusive, while DeepSeek's Diagnosis improves by 4.60 points. These crossed evaluations characterize compatibility along one refinement trajectory; they do not compare PairRoute with an independently trained alternative update method.

\FloatBarrier

%% file: figures/table_grpo_probe.tex
\begin{table}[!htbp]
\centering
\small
\caption{GRPO training-path probe on LongTutor Gold1000. The same checkpoints are evaluated with direct inference and with the final $H_1$ harness. Parentheses report the paired change from G1 to G2; Evidence and Diagnosis are percentages, and Teaching is on a 1--5 scale.}
\label{tab:grpo-probe}
\setlength{\tabcolsep}{4pt}
\begin{tabular*}{0.92\linewidth}{@{\extracolsep{\fill}}lrrrr@{}}
\toprule
\textbf{Checkpoint} & \textbf{Interface} & \textbf{Evidence} & \textbf{Diagnosis} & \textbf{Teaching}\\
\midrule
G1 & Direct & 30.9 & 31.6 & 2.768\\
G2 & Direct & 30.1 & 28.0 & 2.758\\
G1 & RegulaTutor $H_1$ & 47.2 & 49.0 & 2.917\\
G2 & RegulaTutor $H_1$ & \textbf{64.6} & 48.3 & 2.884\\
\midrule
\multicolumn{2}{l}{G1 $\rightarrow$ G2 under $H_1$} & \textcolor{appgreen}{+17.3} & $-0.7$ & $-0.034$\\
\bottomrule
\end{tabular*}
\end{table}